\pdfoutput=1
\documentclass[11pt]{article}

\usepackage{emnlp2021}

\usepackage{times}
\usepackage{latexsym}

\usepackage[T1]{fontenc}
\usepackage[utf8]{inputenc}

\usepackage{microtype}

\usepackage{url}
\usepackage{slashbox}
\usepackage{makecell}
\usepackage{textcomp}
\usepackage{amsmath}
\usepackage{graphicx}
\usepackage{tikz}
\usepackage{arydshln}
\usepackage{array} 
\usepackage{slashbox}
\usepackage{multirow}
 \usepackage{float}
\usetikzlibrary{calc}

\def\imagetop#1{\vtop{\null\hbox{#1}}}
\makeatletter
\newcommand{\thickhline}{%
    \noalign {\ifnum 0=`}\fi \hrule height 1pt
    \futurelet \reserved@a \@xhline
}
\makeatother

\title{Models and Datasets for Cross-Lingual Summarisation}

\author{Laura Perez-Beltrachini \and Mirella Lapata \\
Institute for Language, Cognition and Computation\\
School of Informatics, University of Edinburgh\\
10 Crichton Street, Edinburgh EH8 9AB\\
 \texttt{\{lperez,mlap\}@inf.ed.ac.uk}}

\begin{document}
\maketitle
\begin{abstract}

  We present a cross-lingual summarisation corpus with long documents
  in a source language associated with multi-sentence summaries in a
  target language.  The corpus covers twelve language pairs and
  directions for four European languages, namely Czech, English,
  French and German, and the methodology for its creation can be
  applied to several other languages.  We derive cross-lingual
  document-summary instances from Wikipedia by combining lead
  paragraphs and articles' bodies from language aligned Wikipedia
  titles.  We analyse the proposed cross-lingual summarisation task
  with automatic metrics and validate it with a human study.  To
  illustrate the utility of our dataset we report experiments with
  multi-lingual pre-trained models in supervised, zero- and few-shot,
  and out-of-domain scenarios.

\end{abstract}

\section{Introduction}

\begin{figure*}[t]
\centering
 \begin{tabular}{cc}
  \imagetop{\includegraphics[scale=.45]{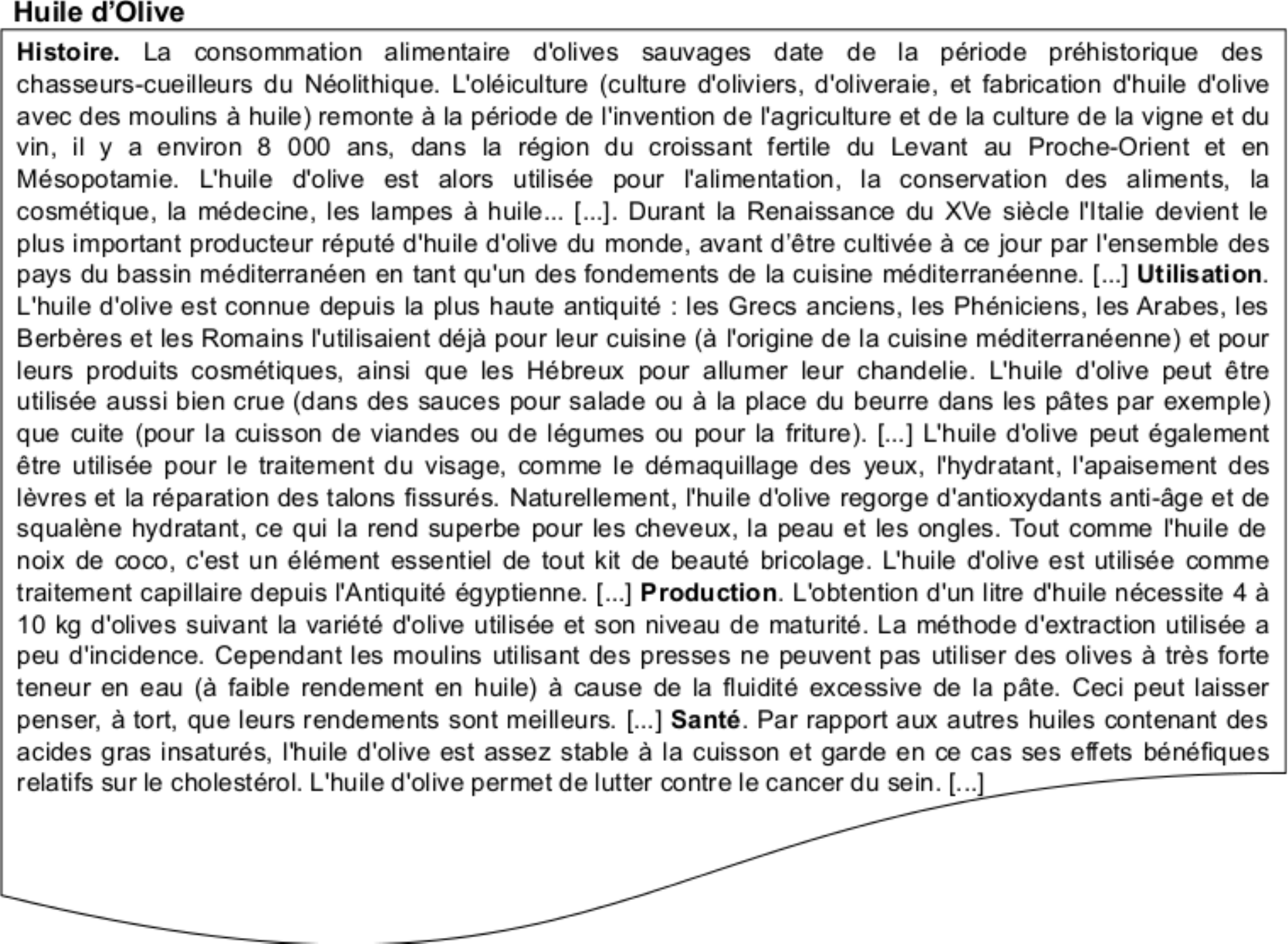}}
  &
 \imagetop{\includegraphics[scale=.45]{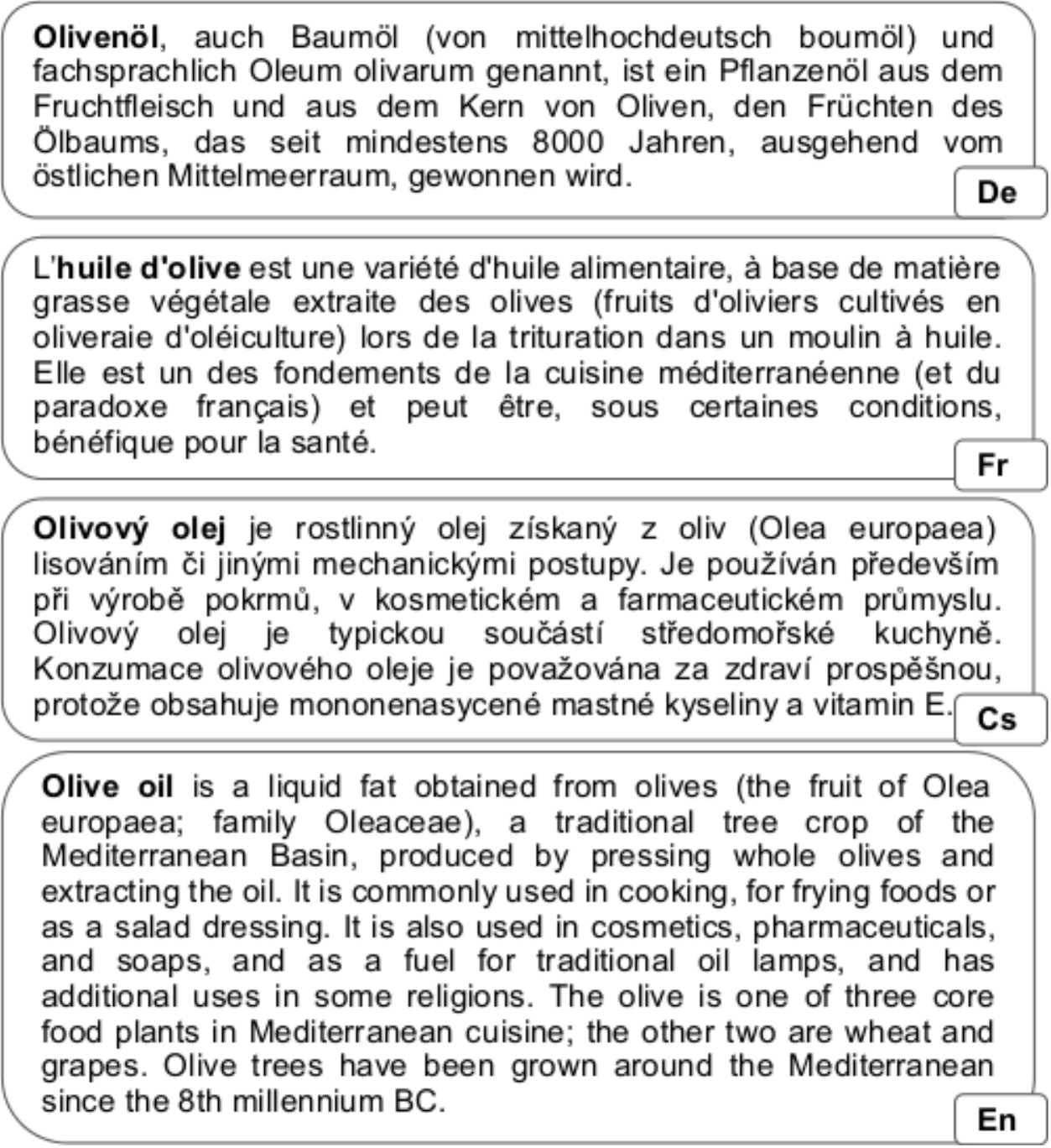}}
 \end{tabular}
 \caption{Example source document in French and target summaries in German, French, Czech and English.}
 \label{fig:clads-example}
 \end{figure*}

Given a document in a source language (e.g., French), cross-lingual
summarisation aims to produce a summary in a different target language
(e.g., English).  The practical benefits of this task are twofold: it
not only provides rapid access to salient content, but also enables
the dissemination of relevant content across speakers of other
languages. For instance, providing summaries of articles from French
or German newspapers to non-French or non-German speakers; or enabling
access to summary descriptions of goods, services, or knowledge
available online in foreign languages.  Figure~\ref{fig:clads-example}
shows an example of an input document in French (left) and its summary
in English and other languages (right).

Recent years have witnessed increased interest in abstractive
summarisation \cite{rush2015neural,zhang2020pegasus} thanks to the
popularity of neural network models and the availability of datasets
\cite{nytcorpus,hermann-nips15,grusky-etal-2018-newsroom} containing
hundreds of thousands of document-summary pairs. Although initial
efforts have overwhelmingly focused on English, more recently, with
the advent of cross-lingual representations
\cite{crossling-ruder.et.al2019} and large pre-trained models
\cite{devlin-etal-2019-bert,mbart}, research on multi-lingual
summarisation (i.e., building monolingual summarisation systems for
different languages) has been gaining momentum
\cite{chi2019crosslingual,scialom2020mlsum}.

While creating large-scale multi-lingual summarisation datasets has
proven feasible \cite{sumeczech-2018,scialom2020mlsum}, at least for
the news domain, cross-lingual datasets are more difficult to obtain.
In contrast to monolingual summarisation, naturally occurring
documents in a source language paired with summaries in different
target languages are rare.  For this reason, existing approaches
either create large-scale synthetic data using back-translation
\cite{zhu2019ncls,cao-etal-2020-jointly}, translate the input
documents \cite{ouyang-etal-2019-robust}, or build document-summary
pairs from social media annotations and crowd-sourcing
\cite{nguyen-daume-iii-2019-global}.  Recent efforts
\cite{ladhak-etal-2020-wikilingua} have been directed at the creation
of a large-scale cross-lingual dataset in the domain of how-to guides.
Despite being a valuable resource, how-to guides are by nature
relatively short documents (391 tokens on average) and their summaries
limited to brief instructional sentences (mostly commands).

 To further push research on cross-lingual summarisation, we propose a
 large dataset with document-summary pairs in four languages: Czech,
 English, French, and German.\footnote{Although we focus on this
   language subset in this paper, we plan to release further languages
   in the future.}  Inspired by past research on monolingual
 descriptive summarisation
 \cite{sauper-barzilay-2009-automatically,zopf-2018-auto,liu2018generating,liu-lapata-2019-hierarchical,perez-beltrachini-etal-2019-generating,wikiasp2021},
 we derive cross-lingual datasets from
 Wikipedia\footnote{\url{https://www.wikipedia.org/}}, which we
 collectively refer to as XWikis.  We exploit Wikipedia's
 Interlanguage links and assume that given any two related Wikipedia
 titles, e.g.,~\textit{Huile d'Olive} (French) and \textit{Olive Oil}
 (English), we can pair the the lead paragraph from one title with the
 body of the other.  We assume that the lead paragraph can stand as
 the summary of the article (see Figure~\ref{fig:clads-example}).  Our
 dataset covers different language pairs and enables different
 summarisation scenarios with respect to: degree of supervision
 (supervised, zero- and few-shot), combination of languages
 (cross-lingual and multi-lingual), and language resources (high- and
 low-resource).

 To illustrate the utility of our dataset we report experiments on
 supervised, zero-shot, few-shot, and out-of-domain cross-lingual
 summarisation.  For the out-of-domain setting, we introduce Voxeurop,
 a cross-lingual news dataset.\footnote{\url{http://voxeurop.eu}. 
 We were given authorisation by Voxeurop SCE publishers 
 \url{https://voxeurop.eu/en/legal-notice-privacy/}} In
 experiments, following recent work \cite{ladhak-etal-2020-wikilingua}, 
 we focus on All-to-English summarisation. In addition to assessing 
 supervised and zero-shot performance of multilingual pre-trained 
 models \cite{mbart,tang2020multilingual}, we also provide a training
 mechanism for few-shot cross-lingual summarisation.\footnote{Code and 
 data are available at \url{https://github.com/lauhaide/clads}.}

\section{The XWikis Corpus}
\label{sec:dataset}

Wikipedia articles are organised into two main parts, a lead section
and a body. For a given Wikipedia title, the lead section provides an
overview conveying salient information, while the body provides
detailed information. Indeed, the body is a long multi-paragraph text
generally structured into sections discussing different aspects of the
Wikipedia title.  We can thus consider the body and lead paragraph as
a document-summary pair. Furthermore, a Wikipedia title can be
associated with Wikipedia articles in various languages also composed
by a lead section and body.  Based on this insight, we propose the
cross-lingual abstractive document summarisation task of generating an
overview summary in a target language $Y$ from a long structured input
document in a source language $X$. Figure~\ref{fig:clads-example}
illustrates this with an example. For the Wikipedia title
\textit{Huile d'Olive} (\textit{Olive Oil}), it shows the French
document on the left and overview summaries in German, French, Czech,
and English on the right.

Below, we describe how our dataset was created, analyse its main
features (Section~\ref{sec:dataset:characterisation}), and present a
human validation study (Section~\ref{sec:dataset:validation}).

\paragraph{Cross-Lingual Summarisation Pairs} 

From a set of Wikipedia titles with articles (i.e.,~lead paragraph and
body) in $N$ languages, we can create ${\frac{N!}{(N-2)!}}$
cross-lingual summarisation sets ${\cal D_{X \rightarrow Y}}$,
considering all possible language pairs and directions.  Data points
(Doc$_{X}$, Sum$_{Y}$) in ${\cal D_{X \rightarrow Y}}$ are created, as
discussed in the previous section, by combining the body of articles
for titles $t_X$ in language~$X$ with the lead paragraph of articles
for corresponding titles $t_Y$ in language~$Y$.  In this work, we
focus on four languages, namely English (en), German (de), French
(fr), and Czech (cs).

To create such summarisation sets ${\cal D_{X \rightarrow Y}}$, we
first use Wikipedia Interlanguage Links to align titles across
languages, i.e., align title $t_X \in X$ with $t_Y \in
Y$.\footnote{\url{https://en.wikipedia.org/wiki/Help:Interlanguage_links}}
Then, from the aligned titles $t_X - t_Y$, we retain those whose
articles permit creating a data point (Doc$_{X}$, Sum$_{Y}$).  In
other words, $t_X$'s article body and $t_Y$'s lead section should obey
the following length restrictions: a) the body should be between 250
and 5,000 tokens long and b) and the lead between 20 and 400
tokens. Table~\ref{tab:xwikis-pairs} shows the number of instances in
each set ${\cal D_{X \rightarrow Y}}$ that we created following this
procedure.

Wikipedia titles exist in different language subsets,
thus, language sets ${\cal D_{X \rightarrow Y}}$ 
will include different sets of titles.
For better comparison in the evaluation of our models,
we would like to have exactly the same set of titles.
To achieve this, we take 7,000 titles in the intersection 
across all language sets. We call this subset
XWikis-parallel and the sets with remaining 
instances XWikis-comparable.

For further details about the data collection process,
see the Appendix~\ref{sec:appendix:dataset}.

\begin{table}[t]
\begin{center}
\begin{tikzpicture}
\node (table) {
{%\small
  \begin{tabular}{@{\hspace{6pt}}c@{\hspace{6pt}}c@{\hspace{6pt}}
  c@{\hspace{6pt}}c@{\hspace{6pt}}c@{}c@{}}
 
\thickhline
 \backslashbox{{\small ${\cal X}$}}{{\small ${\cal Y}$}}  & en & de  & fr & cs \\
en &  & 425,279 & 468,670 & 148,519 \\ 
de &  376,803 & & 252,026 & 109,467 \\ 
fr &  312,408 & 213,425 & & \hspace{0.5em}91,175 \\ 
cs & \hspace{0.5em}64,310 & \hspace{0.5em}53,275 & \hspace{0.5em}51,578 & \\ 
 \thickhline
  \end{tabular}  
 }
};
\end{tikzpicture}
\end{center}
\caption{Total number of document-summary pairs in the XWikis 
corpus considering all language pairs and directions.
Each table cell corresponds to a cross-lingual dataset 
${\cal D_{X \rightarrow Y}}$.}\label{tab:xwikis-pairs}
\end{table}

\paragraph{Monolingual Summarisation Data} A by-product of our data
extraction process is the creation of multi-lingual summarisation
data. Each ${\cal D_{X \rightarrow Y}}$ set has its corresponding
monolingual ${\cal D_{X \rightarrow X}}$ version.  Data points
(Doc$_{X}$, Sum$_{X}$) in ${\cal D_{X \rightarrow X}}$ are created by
combining the body of articles for titles $t_X$ in language~$X$ with
the lead paragraph of articles in the same language~$X$.

\subsection{Features of XWikis Dataset}
\label{sec:dataset:characterisation}

\paragraph{Comparison with Existing Datasets}
Our dataset departs from existing datasets in terms of size,
summarisation task, and potential for extension to additional
languages.  Table~\ref{tab:xwikis-comparison} shows statistics for our
XWikis corpus and existing datasets. Our dataset is larger in terms of
number of document-summary pairs.  WikiLingua
\cite{ladhak-etal-2020-wikilingua} is also larger than previous
datasets, in terms of number of instances, however, the summarisation
task is different.  In XWikis, the input documents are more than twice
as long (average number of tokens).  As for the number of languages,
although in this work we focus on four European ones, the proposed
data creation approach allows to extend the dataset to a large number
of languages including more distant pairs (e.g.,~English-Chinese), as
well as low-resource and understudied languages (e.g.,~Gujarati and Quechua).

\begin{table}[t]
\begin{center}
{
  \begin{tabular}{@{}m{2.5cm}@{\hspace{0.1cm}}r@{\hspace{6pt}}r@{\hspace{6pt}}
  r@{\hspace{6pt}}r@{}r@{}}
 
\thickhline
 Dataset  &     Lang &  Pairs  & SumL &  DocL \\\thickhline
MultiLing’13 &  40 &      30 &  185 & 4,111 \\ 
MultiLing’15 &  38 &      30 &  233 & 4,946 \\ 
Global Voices & 15 &     229 &   51 &   359 \\ 
WikiLingua &    18 &  45,783 &   39 &   391 \\ 
XWikis (comp.) & 4 & 213,911 &   77 &   945 \\
XWikis (para.) & 4 &   7,000 &   76 &   972 \\
 \thickhline
  \end{tabular}  
}
\end{center}
\caption{Number of languages (Lang), average number of document-summary 
pairs (Pairs), average summary (SumL) and document (DocL) length
in terms of number of tokens.}\label{tab:xwikis-comparison}
\end{table}

\paragraph{Summarisation Task}
We carry out a detailed analysis of our XWikis corpus to characterise
the summarisation task it represents and assess the validity of the
created summarisation data points (Doc$_{X}$, Sum$_{Y}$). In the first
instance, we do this through automatic metrics. Since metrics that are
based on token overlap \cite{grusky-etal-2018-newsroom,narayan2018don}
cannot be directly applied to our cross-lingual data, we carry out
some automatic analysis on the monolingual version of the corpus
instead, i.e.,~(Doc$_{X}$, Sum$_{X}$) instances.  We first validate
the assumption that the lead paragraph can serve as a summary for the
article body.  Table~\ref{tab:sum-task} provides statistics per
language pair, for XWikis-comparable\footnote{We compute statistics
  and metrics on the monolingual subsets from de-en, fr-en, and
  cs-en.}, and averaged over all language pairs for XWikis-parallel.

\textit{Size.}  The top part of Table~\ref{tab:sum-task} provides an
overview of the summarisation task in terms of size.  The documents
are long, with an overall average of 952 tokens, 40 sentences (note
that sentence length is thus \texttildelow23 tokens) and 6 sections.
Such lengthy documents are challenging for current neural
summarisation models which struggle to represent multi-paragraph text;
most approaches rely on an initial separate extractive step
\cite{liu2018generating,liu-lapata-2019-hierarchical, perez-beltrachini-etal-2019-generating}.  Each section
describes a different aspect of its related Wikipedia title
\cite{wikiasp2021}.  We analyse the average number of sections per
document as a proxy for the complexity of the content selection
sub-task.  A summariser will need to learn which aspects are
summary-worthy and extract content from different sections in the
input document.  Summaries are also long with 60 tokens and 3
sentences on average.

\begin{table}[t!]
\begin{center}
{
\begin{tabular}{@{}m{2.5cm}@{\hspace{0.1cm}}c@{\hspace{5pt}}c@{\hspace{5pt}}c@{}c@{}c@{}}
\thickhline
  &  \multicolumn{3}{c}{XWikis (comp)} & XWikis  \\
  &  de & fr & cs & (para) \\
\thickhline
Words/Doc  &   906      & 1040    & 890     & 972 \\
Sents/Doc &  41  & 38 & 42 & 42 \\ %Also have statistics of average sentence length
Sections/Doc & 5 & 7 & 6 & 6\\
Words/Sum &   56 & 59 & 65 & 61 \\
Sents/Sum & 3  & 2  & 3  & 3 \\ %Also have statistics of average sentence length
Aspects  & 253,425 & 248,561 &  65,151 & 9,283\\%[1.1ex] % only <h2> sections
Coverage    & 65.53 & 72.23 & 55.97 & 65.41 \\
Density     &  1.23 &  1.51 &  0.99 &  1.23 \\
Compression & 17.44 & 20.16 & 15.12 & 18.35 \\
\% new unigrams    & 33.30 & 26.77 & 42.29 & 33.25 \\
\hspace{1.2cm}bigrams     & 80.70 & 73.19 & 85.17 & 79.51 \\
\hspace{1.2cm}trigrams    & 93.60 & 90.25 & 95.19 & 93.17 \\
\hspace{1.2cm}4-grams     & 97.98 & 95.68 & 97.98 & 97.11 \\
\textsc{Lead} & 19.09 & 23.51 & 20.21 & 20.88 \\
\textsc{Ext-Oracle} & 24.59 & 28.38 & 24.25 & 25.95 \\
\thickhline
\end{tabular}  
}
\end{center}
\caption{XWikis statistics (number of words and sentences per document (/Doc)
  and summary (/Sum)) and task characterisation
  metrics.}\label{tab:sum-task} 
\end{table}

\textit{Content Diversity.} To assess the diversity of content in
the corpus, we report the number of distinct top level section titles
as an approximation (without doing any normalisation) of aspects
discussed \cite{wikiasp2021}.  These high numbers, together with the
average number of sections per document, confirm that our dataset
represents multi-topic content.

\textit{Level of Abstraction.}  To characterise the summarisation task
in terms of level of abstraction, we analyse content overlap of
document-summary pairs using automatic metrics
\cite{grusky-etal-2018-newsroom,narayan2018don} and then evaluate the
performance of two extractive summarisation
approaches.\footnote{Extractive methods were run on validation
  splits.} When the summarisation task is extractive in nature
(i.e.,~the summaries copy text spans from the input document),
extractive methods ought to perform well.

The set of automatic metrics proposed in
\citet{grusky-etal-2018-newsroom}, indicates the extent to which a
summary is composed by textual fragments from the input document,
i.e., extractive fragments. \textit{Coverage}, measures the average
number of tokens in the summary that are part of an extractive
fragment; \textit{Density}, indicates the average length of the set of
extractive fragments. As shown in Table~\ref{tab:sum-task},
\textit{Coverage} is high, specially for de and fr sets, while
\textit{Density} is quite low. This indicates that the summaries
overlap in content with the input documents but not with the same
phrases.  Although summaries are not short, the compression ratio is
high given the size of the input documents.  This highlights the
rather extreme content selection and aggregation imposed by the
summarisation task.  The second set of metrics proposed in
\citet{narayan2018don}, measures the percentage of new n-grams
appearing in the summary (i.e., not seen in the input document), and
shows a similar trend. The percentage of novel unigrams is low but
increases sharply for higher ngrams.

The last two rows in Table~\ref{tab:sum-task} report ROUGE-L
for two extractive methods.  \textsc{Lead} creates a summary by
copying the first $K$~tokens of the input document, where $K$~is the
size of the reference and performs well when the summarisation task is
biased to content appearing in the first sentences of the document.
\textsc{Ext-Oracle} selects a subset of sentences that maximize ROUGE-2
\cite{lin-2004-rouge} with respect to the reference summaries
\cite{AAAI1714636,narayan2018don} and performs well when the
summarisation task is mostly extractive. As we can see, \textsc{Lead}
is well below \textsc{Ext-Oracle} (\texttildelow 4~ROUGE-L points on
average), indicating no lead bias (i.e.,~summary-worthy content is not
in the beginning of the document).  \textsc{Ext-Oracle} performs
better, however, considering the high levels of \textit{Coverage}, it
does not seem to cover all salient content. This indicates that
important content is scattered across the document in different
sentences (not all of which are selected by \textsc{Ext-Oracle}) and
that phrasing is different (see jump from \% of novel unigrams to
bigrams). The French subset, has the highest \textit{Coverage}
(conversely the lower \% of novel unigrams), and thus is more amenable
to the extractive methods.

\subsection{Validation through Human Evaluation}
\label{sec:dataset:validation}

To further complement automatic evaluation, we carried out a human
evaluation study to assess the quality of cross-lingual data
instances (Doc$_{X}$, Sum$_{Y}$).  In other words, we validate the
assumption that given a pair of aligned titles $t_X-t_Y$, the lead
paragraph in language $Y$ is a valid overview summary of the document
body in language $X$.

As this evaluation requires bilingual judges, we selected three
language pairs, namely ${\cal{D}}_{de \rightarrow en}$, ${\cal{D}}_{fr
  \rightarrow en}$ and ${\cal{D}}_{cs \rightarrow en}$ and recruited
three judges per pair, i.e., bilingual in German-English,
French-English, and Czech-English. We selected 20 data instances from
each set and asked participants to give an overall judgement of summary
adequacy. Specifically, they were asked to provide a
\texttt{yes}/\texttt{no} answer to the question \textit{Does the
  summary provide a general overview of the Wikipedia title?}. In
addition, we elicited more fine-grained judgments by asking
participants to ascertain for each sentence in the summary whether it
was supported by the document. We elicited \texttt{yes}/\texttt{no}
answers to the question \textit{Does the sentence contain facts that
  are supported by the document?}. We expect judges to answer
\texttt{no} when the content of a sentence is not discussed in the
document and \texttt{yes} otherwise.

\begin{table}[t]
\begin{center}
{%\footnotesize
  \begin{tabular}{m{2cm}@{\hspace{0.05cm}}c@{\hspace{6pt}}c@{\hspace{6pt}}
 c@{\hspace{6pt}}c@{\hspace{6pt}}c@{}}
 
\thickhline
 Dataset  &    de $\rightarrow$ en & fr $\rightarrow$ en &  cs $\rightarrow$ en \\[1.1ex]
\thickhline
%Three annotators
 Overall  & 71.7\% & 96.6\% & 73.3\% \\
 Sentence & 66.2\% & 77.4\% &  60.5\% \\ 
\thickhline 
  \end{tabular}  
}
\end{center}
\vspace{-0.5em}
\caption{Proportion of \texttt{yes} answers given to questions of
  Overall  summary and Sentence adequacy. Judgments elicited for  cross-lingual
  document-summary pairs in three languages.}\label{tab:human-valid}
\end{table} 

Table~\ref{tab:human-valid} shows the proportion of \texttt{yes}
answers given by our judges for the three language pairs. Overall,
judges view the summary as an acceptable overview of the Wikipedia
title and its document. The same picture emerges when considering the
more fine-grained sentence-based judgments.  66.2\% of the summary
sentences are supported by the document in the German-English pair,
77.4\% for French-English, and 60.5\% for Czech-English. We also used
Fleiss's Kappa to establish inter-annotator agreement between our
judges. This was 0.48 for German-English speakers, 0.55 for
French-English, and 0.59~for Czech-English.

\section{All-to-English Summarisation}

\subsection{Task} 

Following previous work \cite{ladhak-etal-2020-wikilingua}, the
specific cross-lingual task that we address is generating English
summaries from input documents in different (source) languages.  In
the context of cross-lingual summarisation, we assume that a)~we have
enough data to train a monolingual summarizer in a source language; b)
we want to port this summarizer to a different target language without
additional data (\emph{zero-shot}) or a handful of training examples
(\emph{few-shot}); and c)~the representations learnt by the
monolingual summarizer to carry out the task, i.e., select relevant
content and organise it in a short coherent text, should transfer or
adapt to the cross-lingual summarisation task.  The main challenges in
this setting are understanding the input documents in a new language
which may have new relevance clues and translating them into the
target language.

Specifically, we assume we have access to monolingual English data
(Doc$_{en}$, Sum$_{en}$) to learn an English summariser, and we study
the zero- and few-shot cross-lingual scenarios when the input to this
model is in a language other than English (i.e.,~German, French, and
Czech). We further exploit the fact that our XWikis corpus allows us
to learn cross-lingual summarisation models in a fully supervised
setting, and establish comparisons against models with weaker
supervision signals. Our fully supervised models follow
state-of-the-art approaches based on Transformers and pre-training
\cite{liu-lapata-2019-text,lewis-etal-2020-bart}. 
We simulate zero- and few- shot scenarios by
considering subsets of the available data instances.

\subsection{Approach} 
\label{sec:allToEnApp}

We formalise cross-lingual abstractive summarisation as follows. Given
input document Doc$_{X}$ in language~$X$ represented as a sequence of
tokens $x=(x_1 \cdots x_{|x|})$, our task is to generate Sum$_Y$ in
language~$Y$.  The target summary is also represented as sequence of
tokens $y=(y_1 \cdots y_{|y|})$ and generated token-by-token
conditioning on $x$ by a summarisation model $p_{\theta}$ as
$\prod_{t=1}^{|y|} p_{\theta}(y_t|y_{1..t-1}, x)$.

Our summarisation model is based on m\textsc{Bart50}
\cite{tang2020multilingual}, a pre-trained multi-lingual
sequence-to-sequence model.  m\textsc{Bart50}
\cite{tang2020multilingual} is the result of fine-tuning
m\textsc{Bart} \cite{mbart} with a multi-lingual machine translation
objective (i.e.,~fine-tuning with several language pairs at the same
time).  The fine-tuning process extends the number of languages
from~25 to~50.  \textsc{Bart} \cite{mbart} follows a Transformer
encoder-decoder architecture \cite{vaswani2017attention}. It was
trained on a collection of monolingual documents in 25 different
languages to reconstruct noised input sequences which were obtained by
replacing spans of text with a \textit{mask} token or permuting the
order of sentences in the input.

Although pre-trained models like m\textsc{Bart50} provide
multi-lingual representations for language understanding and
generation, they require adjustments in order to be useful for
abstractive summarisation.  Given a training dataset ${\cal{D}}$ with
document-summary instances $\{x_n, y_n\}_{n=1}^{|{\cal{D}}|}$ starting
from a model with parameters $\theta$ given by m\textsc{mBart50}, we
fine-tune to minimise the negative log likelihood on the training
dataset, \mbox{${\cal{L}}_{NLL}=- \frac{1}{|{\cal{D}}|}
  \sum_{n=1}^{|{\cal{D}}|} \text{log } p_{\theta}(y_n|x_n)$}.  If
${\cal{D}}$ is instantiated by a cross-lingual dataset
(i.e.,~${\cal{D}}_{X \rightarrow Y}$) we directly fine-tune on the
target cross-lingual task.  However, in our zero and few-shot
settings we only have monolingual summarisation data available. We
therefore assume~${\cal{D}}$ to be an English monolingual set
(i.e.,~${\cal{D}}_{en \rightarrow en}$).

In the \emph{zero-shot} scenario, a monolingual summariser English
summariser is used for cross-lingual summarisation and we assume that
the parameters of the English model will be shared to a certain extent
across languages \cite{chi-etal-2020-finding}.  In the \emph{few-shot}
scenario, we assume that in addition to monolingual summarisation
data, we also have access to a small dataset $S_{X \rightarrow en}$
with cross-lingual summarisation examples. Although it might be
possible to curate cross-lingual summaries for a small number of
examples, using these in practice for additional model adaptation can
be challenging. In this work propose an approach reminiscent of the
few-shot Model Agnostic Meta-Learning (MAML) algorithm
\cite{pmlr-v70-finn17a}.

MAML is an optimisation-based learning-to-learn algorithm which
involves meta-training and meta-testing phases.  Meta-training
encourages learning representations which are useful across a set of
different tasks and can be easily adapted, i.e., with a few data
instances and a few parameter updates, to an unseen task during
meta-testing.  More concretely, meta-training consists of nested
optimisation iterations: inner iterations take the (meta) model
parameters~$\theta_{meta}$ and compute for each task~${\cal T}_i$ a
new set of parameters~$\theta_i$. In the outer iteration, the (meta)
model parameters are updated according to the sum of each task~${\cal
  T}_i$ loss on task-specific parameters $\theta_i$.\footnote{ A
  simplified version, First-Order MAML, updates the (meta) model
  parameters directly with the derivative of the last inner loop
  gradient update \cite{pmlr-v70-finn17a}.}  At test time, the (meta)
model parameters can be adapted to a new task with one learning step
using the small dataset associated with the new task.

We assume that the multi-lingual and MT pre-training of mBART50 (and
mBART) are a form of meta-training involving several language tasks
which learn shared representations across different languages.  We
then adapt the English monolingual summariser to the cross-lingual
task ${\cal T}_{{X \rightarrow en}}$ with a small set of instances $S_{X
  \rightarrow en}$.  We perform a single outer loop iteration and
instead of taking a copy of the (meta) parameters and updating them
after the inner loop, we combine the support set with a monolingual
sample of similar size.  We call this method light-weight First Order
MAML (LF-MAML).

We also observe that in a real-world scenario, in addition to the
small set with cross-lingual examples $S_{X \rightarrow en}$, there
may exist documents in the source language Doc$_X$ without
corresponding summaries in English. To further train the model with
additional unlabelled data, we apply a Cross-View Training technique
(CVT; \citealt{Clark2018SemiSupervisedSM}).  We exploit the fact that
our fine-tuning does not start from scratch but rather from a
pre-trained model which already generates output sequences of at least
minimal quality.  We augment the set of document summary pairs ${x,y}$
in $S_{X \rightarrow en}$ with instances ${\hat{x},\hat{y}}$ where
$\hat{y}$ is generated by the current model and $\hat{x}$ is a
different view of $x$. We cheaply create different views from
input~$x$ by taking different layers from the encoder.

\section{Experimental Setup}

\paragraph{Datasets and Splits}
We work with the ${\cal{D}}_{de \rightarrow en}$, ${\cal{D}}_{fr
  \rightarrow en}$, and ${\cal{D}}_{cs \rightarrow en}$ directions of
our XWikis corpus (i.e., first column in Table~\ref{tab:xwikis-pairs})
and evaluate model performance on the XWikis-parallel set. We split
XWikis-comparable into training (95\%) and validation (5\%) sets. 

To train an English monolingual summariser, we created a monolingual
dataset ${\cal{D}}_{en \rightarrow en}$ following the procedure
described in Section~\ref{sec:dataset} (lead paragraph and body of
Wikipedia articles). We selected a set of Wikipedia titles disjoint
from those in our XWikis corpus. This dataset has 300,000 instances
with~90/5/5 percent of instances in training/validation/test
subsets. It follows similar characteristics to the data in our XWikis
corpus with an average document and summary length of 884 and 70
tokens, respectively.

\paragraph{Paragraph Extraction} 
To deal with very long documents, we carry out an initial extractive
step \cite{liu2018generating,liu-lapata-2019-hierarchical}.
Specifically, we rank document paragraphs (represented as vectors of
their tf-idf values) using \textsc{LexRank} \cite{erkan2004lexrank}
and then select the top ranked paragraphs up to a budget of 600
tokens. Table~\ref{tab:xwikis-stats} reports ROUGE-L recall of the
input against the reference summary (note that to measure this we take
the monolingual summary associated with the document rather than the
cross-lingual one).  As can be seen, the extractive step reduces the
document to a manageable size without sacrificing too much content.
Note that after ranking, selected paragraphs are kept in their
original position to avoid creating a bias towards important
information coming at the beginning of the input sequence.

\begin{table}[t]
\begin{center}
{
\begin{tabular}{m{2cm}@{\hspace{0.1cm}}c@{\hspace{9pt}}c@{}c@{}c@{}}
\thickhline
 & All & 800 & ParaLexRank.600 \\\thickhline
${\cal{D}}_{en \rightarrow en}$ & 55.16 & 51.83 & 51.88 \\
${\cal{D}}_{de \rightarrow en}$ & 52.05 & 48.64 & 48.60 \\   
${\cal{D}}_{fr \rightarrow en}$ & 56.05 & 51.78 & 51.86 \\
${\cal{D}}_{cs \rightarrow en}$ & 53.37 & 50.20 & 50.47\\
\thickhline
\end{tabular}  
}
\end{center}
\caption{ROUGE-L recall for source document against 
  reference monolingual summary computed against all input tokens
  (All),  the first 800 tokens and  the 600 tokens 
  extracted with paragraph-based \textsc{LexRank}.}\label{tab:xwikis-stats}
\end{table}

\paragraph{Out of Domain Data} To evaluate the robustness of
cross-lingual models on non-Wikipedia text, we created an out of
domain dataset from the European news site Voxeurop.  This site
contains news articles composed of a summary section (with
multi-sentence summaries) and a body written and translated into
several languages by professional journalists and translators. We
extracted from this site 2,666 summary-article pairs in German,
French, Czech, and English. The average document length in tokens
is~842 and the summary length~42. We used 2,000 instances for
evaluation and reserved the rest for model adaptation.

\subsection{Models}

We evaluated a range of extractive and abstractive summarisation
models detailed below. In cases where translation is required we used
the Google~API.\footnote{Translation was supported by Google Cloud  Platform credits.}

\paragraph{Extractive} We applied extractive approaches on the source
documents.  Extracted sentences were the translated into English to
create a summary in the target language.

\begin{enumerate}\itemsep0pt
\item \textsc{Ext-Oracle} This extractive approach builds summaries by
  greedily selecting sentences from the input that together maximize
  ROUGE-2 against the reference summary. We implemented this upper
  bound following \citet{AAAI1714636}'s procedure. For datasets
  ${\cal{D}}_{X \rightarrow en}$, we take the monolingual summary
  associated to the input document as a proxy for ROUGE-based
  selection.

\item \textsc{Lead} The first~$K$ tokens from the input document are
  selected where~$K$ is the length of the reference summary.

\item \textsc{LexRank} This approach uses tf-idf graph-based sentence
  ranking \cite{erkan2004lexrank} to select sentences from the input
  and then takes first $K$~tokens (where~$K$ is the length of the
  reference summary).
\end{enumerate}

\paragraph{Supervised}
We fine-tuned three separate models based on m\textsc{Bart}
\cite{mbart} and m\textsc{Bart50} \cite{tang2020multilingual} in a
supervised fashion on the three cross-lingual datasets (${\cal{D}}_{de
  \rightarrow en}$, ${\cal{D}}_{fr \rightarrow en}$, and
${\cal{D}}_{cs \rightarrow en}$).  This provides an upper-bound on
achievable performance.  Additionally, we trained an English
summariser on the separate English dataset ${\cal{D}}_{en \rightarrow
  en}$ (described in the previous section) for our zero and few-shot
scenarios.

\paragraph{Translated} This is a translate and summarise pipeline
approach. We first translate the input documents Doc$_{de}$,
Doc$_{fr}$, and Doc$_{cs}$ into English and then apply a monolingual
English summariser.

\vspace{-0.3em}
\paragraph{Zero-Shot} A monolingual English summariser is directly
applied to summarise Doc$_{de}$, Doc$_{fr}$, and Doc$_{cs}$ documents
into English. We fine-tune the entire network except the embedding
layer. We report experiments with m\textsc{Bart50} (and
m\textsc{Bart}).

\vspace{-0.3em}
\paragraph{Few-Shot} These models are based on fine-tuned monolingual
English summarisers subsequently adapted to the cross-lingual task
with a small set of examples $S_{X \rightarrow en}$. We present
experiments with m\textsc{Bart} and m\textsc{Bart50} pre-trained
models. We evaluate three few-shot variants (see
Section~\ref{sec:allToEnApp}). \mbox{LF-MAML} is the light-weight
First Order MAML version, FT is a fine-tuned version where only
cross-attention and layer normalisation layers are fine-tuned, and CVT
incorporates additional unlabelled instances into the adaptation
step. We also consider two settings with $|S_{X \rightarrow en}|$
being 300 and 1,000 few instances.  Note that in each case we take
$1/3$ for validation, and the rest for training.  For CVT, we generate
two views, $\hat{x}_m$ and $\hat{x}_u$, for each input document $x$ in
$S_{X \rightarrow en}$ by taking a middle encoder representation
($\hat{x}_m$ the hidden states at layer 6) and another by taking an
upper encoder representation ($\hat{x}_u$ the hidden states at layer
11).  Intuitively, these provide different levels of abstraction from
the input document.

\section{Results and Analysis}
In this section we discuss our cross-lingual summarisation results
(Table~\ref{tab:XWikisSupervisedTestROUGEL} and
Table~\ref{tab:XNewsROUGEL}). We provide examples of model output (and
additional experiments) in Appendix~\ref{sec:appendix:results}.

\begin{table}[t]
\centering
{
  \begin{tabular}{@{}c@{\hspace{0.2cm}}c@{\hspace{0.2cm}}m{2.6cm}@{\hspace{0.1cm}}c@{\hspace{6pt}}c@{\hspace{6pt}}
  c@{\hspace{6pt}}c@{}c@{}c@{}}
 
\thickhline
  & & & en & de-en & fr-en  & cs-en  \\
\thickhline

 \multicolumn{3}{l}{\textsc{Ext-Oracle}} & 31.33 & 23.75 & 25.01 & 25.09 \\
 \multicolumn{3}{l}{\textsc{Lead}} & 25.45 & 24.95 & 24.74 & 24.35 \\
 \multicolumn{3}{l}{\textsc{LexRank}} & 25.23 & 24.22 & 24.33 & 23.68 \\%[1.2ex]
 
 \hline
  \parbox[t]{2mm}{\multirow{4}{*}{\rotatebox[origin=m]{90}{m\textsc{Bart}\hspace*{1.1cm}}}} &
 \multicolumn{2}{l}{Supervised} & 31.62 & 32.37 & 32.18 & 32.84 \\[1ex] % decode >> all-en=mlb150.ml50.lenpen2 en=mlb150.ml50.lenpen1
 & \multicolumn{2}{l}{Translated} & --- & 30.69 & 30.63 & 30.39 \\[1ex]
 & \multicolumn{2}{l}{Zero} & --- & 30.10 & 29.78 & 28.64 \\[1ex] %decode >> mlb150.ml50.lenpen2
 
  & \parbox[t]{2mm}{\multirow{4}{*}{\rotatebox[origin=m]{90}{Few}}}
    & 300 LF-MAML    & --- & 30.84 & 30.44 & 30.15 \\
  & & 300 FT         & --- & 31.06 & 30.39 & 30.36 \\
  & & 300 CVT        & --- & 30.40 & 30.12 & 29.39 \\
  & & 1K  LF-MAML    & --- & 31.19 & 30.77 & 31.02 \\[1.2ex]
  
 \hline
 \parbox[t]{2mm}{\multirow{9}{*}{\rotatebox[origin=m]{90}{\hspace*{.6cm}m\textsc{Bart50}}}} 
  & \multicolumn{2}{l}{Supervised}  & 32.53 & 32.95 & 31.84 & 33.72 \\[1ex]
  
  & \multicolumn{2}{l}{Translated} & --- & 31.53 & 31.35 & 31.25 \\[1ex]
  
  & \multicolumn{2}{l}{Zero} & --- & 31.70 & 30.97 & 31.14 \\[1ex] 
  
  & \parbox[t]{2mm}{\multirow{4}{*}{\rotatebox[origin=m]{90}{Few}}}
    & 300 LF-MAML    & --- & 31.96 & 31.17 & 31.73 \\
  & & 300 FT         & --- & 31.77 & 31.39 & 31.67 \\
  & & 300 CVT        & --- & 31.77 & 31.08 & 31.91 \\
  & & 1K  LF-MAML    & --- & 32.01 & 31.46 & 32.00 \\
 
\thickhline
  \end{tabular}  
 }
 \caption{ROUGE-L F1 $X \rightarrow en$ XWikis test sets.}\label{tab:XWikisSupervisedTestROUGEL}
\end{table}

 \begin{table}[t]
 \centering
 {
  \begin{tabular}{@{}c@{\hspace{0.2cm}}c@{\hspace{0.2cm}}m{2.6cm}@{\hspace{0.1cm}}c@{\hspace{6pt}}c@{\hspace{6pt}}
  c@{\hspace{6pt}}c@{}c@{}c@{}}
  
 \thickhline
   &&& en & de-en & fr-en & cs-en  \\
 \thickhline
     
 \multicolumn{3}{l}{\textsc{Ext-Oracle}}  & 20.83 & 17.81 & 17.90 & 17.63 \\
 \multicolumn{3}{l}{\textsc{Lead}}        & 17.17 & 17.13 & 16.61 & 17.07 \\
 \multicolumn{3}{l}{\textsc{LexRank}}     & 16.65 & 16.32 & 16.32 & 16.48 \\[1.2ex]
 
 \hline
 \parbox[t]{2mm}{\multirow{3}{*}{\rotatebox[origin=m]{90}{m\textsc{Bart}\vspace{1ex}}}} 
& \multicolumn{2}{l}{Zero} & 21.68 & 19.54 & 19.49 & 18.92 \\
& \parbox[t]{2mm}{\multirow{2}{*}{\rotatebox[origin=m]{90}{Few}}} 
& 300 LF-MAML &  ---     & 22.32 & 22.42 & 22.26 \\
& & 300 FT      &  ---   & 21.86 & 21.74 & 21.72 \\\hline

\parbox[t]{2mm}{\multirow{3}{*}{\rotatebox[origin=m]{90}{m\textsc{Bart50}}}} 
& & \\[-0.9ex]
& \multicolumn{2}{l}{Zero} & 21.28 & 21.04 & 20.66 & 21.30 \\ 
& \parbox[t]{2mm}{\multirow{2}{*}{\rotatebox[origin=m]{90}{Few}}} 
& 300 LF-MAML &  ---     & 21.87 & 21.90 & 22.11 \\
& & 300 FT      &  ---   & 21.79 & 21.53 & 21.95 \\

 \thickhline
   \end{tabular}  
  }
  \caption{ROUGE-L F1 $X \rightarrow en$ Voxeurop test sets.}\label{tab:XNewsROUGEL}
 \end{table} 

 \vspace{-0.3em}
 \paragraph{Does Zero-Shot Work?} Zero-shot (with a monolingual
 English summariser) grasps the gist of the document, and some
 representations are indeed transferred. Despite the summariser being
 learnt on monolingual English data, when presented with documents in
 other languages (i.e., German/French/Czech) it manages to produce a
 summary which, according to ROUGE, is better than extractive
 baselines (including \textsc{Ext-Oracle}).  However, across
 languages, zero-shot results are below the \emph{Supervised}
 upper-bound (see second block in
 Table~\ref{tab:XWikisSupervisedTestROUGEL}). This gap is highest when
 summarizing from Czech to English.
 
\vspace{-0.3em}
 \paragraph{Can we Beat Machine Translation?} In agreement with previous work
 \cite{ladhak-etal-2020-wikilingua}, we find that \textit{Supervised}
 models are better than \textit{Translated} ones.  \textit{Zero}
 versions with m\textsc{BART50} perform slightly below
 \textit{Translated}, except for German-to-English (this is more
 surprising for m\textsc{BART} which has not seen any cross-language
 links during pre-training).  Interestingly, \textit{Few} with
 m\textsc{BART50} and 300 training instances achieves comparable
 performance, which indicates that the summariser can improve on the
 new cross-lingual task by seeing only a few examples.  We observe a
 similar trend for m\textsc{BART} even though it never sees any
 cross-lingual examples during pre-training.

\vspace{-0.3em}
 \paragraph{Which Few-Shot Model is Better?} \mbox{FL-MAML} performs
 well across languages both in the 300 and 1K settings. Indeed, in
 this last configuration it beats \textit{Translated} and gets closer
 to \textit{Supervised} using a relatively small training set (\textasciitilde 600
 instances --- the rest is used for validation). The performance of FT
 and CVT variants varies depending on the language. FT (which only
 fine-tunes cross-attention) helps when summarizing from French
 whereas CVT helps when summarizing from Czech. The latter model
 benefits from potentially noisier unlabelled instances.

\vspace{-0.3em}
\paragraph{Is Out-of-Domain Summarisation Feasible?}
Table~\ref{tab:XNewsROUGEL} shows the performance of a monolingual
English summariser (trained on XWikis) and tested on the Voxeurop
dataset. There is indeed a penalty for domain shift by
approximately~10 ROUGE points (compare row Zero in
Table~\ref{tab:XNewsROUGEL} with rows Supervised/Zero in
Table~\ref{tab:XWikisSupervisedTestROUGEL}). Overall,
\textit{Few}-shot manages to improve upon \emph{Zero}-shot, even
though the few training instances come from a more distant
distribution than the one used to pre-train the monolingual summariser
(i.e.,~different~genres).

\vspace{-0.3em}
\paragraph{Which Pre-trained Model?}  Our experiments identify
m\textsc{BART} as the weakest pre-trained model, reporting lower ROUGE
scores across languages, domains, and training settings (e.g.,
supervised, zero- and few-shot). m\textsc{BART50} benefits from
fine-tuning on machine translation and this knowledge is useful
to our summarisation task.

\vspace{-0.3em}
\paragraph{Are there  Differences between Languages?}
In the XWikis corpus (and mostly with m\textsc{BART}) Czech-to-English
has the lowest performance. However, this gap disappears when applying
\mbox{\textit{Few}-shot} variants to the summarisation task. In
Voxeurop, there are no discernible differences amongst language pairs;
this is probably due to the fact that document-summary pairs are
translations across languages.

\paragraph{How Hard is Cross-lingual Summarisation?} The task is very
challenging! XWikis documents are long, and summarisation models must
be able to represent multi-paragraph text adequately and isolate
important content which is interspersed through the document. This
difficulty is further compounded by the translation of content in
different languages and the need for models to abstract, rephrase, and
aggregate information. Our results in
Tables~\ref{tab:XWikisSupervisedTestROUGEL} and~\ref{tab:XNewsROUGEL} show
that there is plenty of room for improvement.

\section{Conclusion}

We presented a new summarisation dataset in four languages (German,
French, Czech, and English) which we hope will be a valuable resource
for cross-lingual and monolingual summarisation. We evaluated a wide
range of models on the cross-lingual summarisation task, including
zero- and few- shot variants some of which show promising results.
Future work directions are many and varied. We would like to further
investigate MAML variants for few-shot summarisation, and expand on
document views for CVT (e.g., by looking at semantic roles and
discourse relations).

\paragraph{Acknowledgments} We thank the anonymous reviewers for their
feedback. We  also  thank  Yumo Xu for useful discussions about 
the models. We are extremely grateful to our bilingual annotators
and to Voxeurop SCE publishers.
We gratefully acknowledge the support of the European
Research Council (award number 681760).

\bibliography{clads}
\bibliographystyle{acl_natbib}

\clearpage

\appendix

\section{The XWikis Corpus}
\label{sec:appendix:dataset}

\paragraph{Dataset Creation}

Our corpus was created with English, German, French and Czech
Wikipedia dumps from June
2020.\footnote{\url{https://dumps.wikimedia.org}} We adapted
Wikiextractor \cite{Wikiextractor2015} to obtain the lead section and
body of Wikipedia articles. We preserved the structure of the input
document, and section mark-ups were kept (e.g., \verb|<h2>|).  We used
a dump of the same date for the table containing the Wikipedia
Interlanguage
Links.\footnote{\url{https://en.wikipedia.org/wiki/Help:Interlanguage_links}}
We performed text normalisation (a variant of NFKC normalization) with
 sentence-piece \cite{kudo-richardson-2018-sentencepiece}.

\begin{figure*}[t]
\centering
   \begin{tabular}{cc}
   \imagetop{\includegraphics[scale=.45]{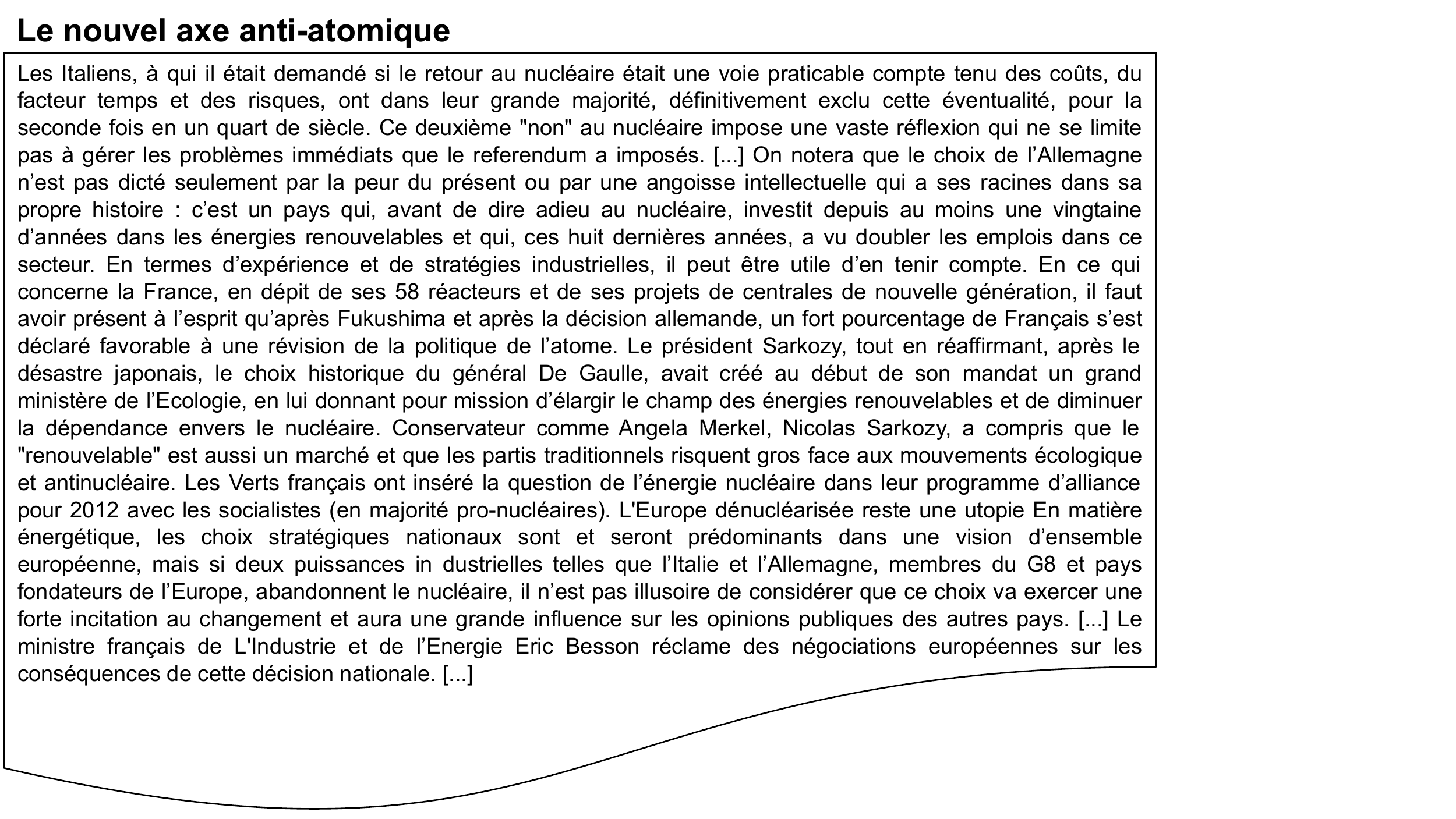}}
   &
   \hspace{-2.4cm}\imagetop{\includegraphics[scale=.5]{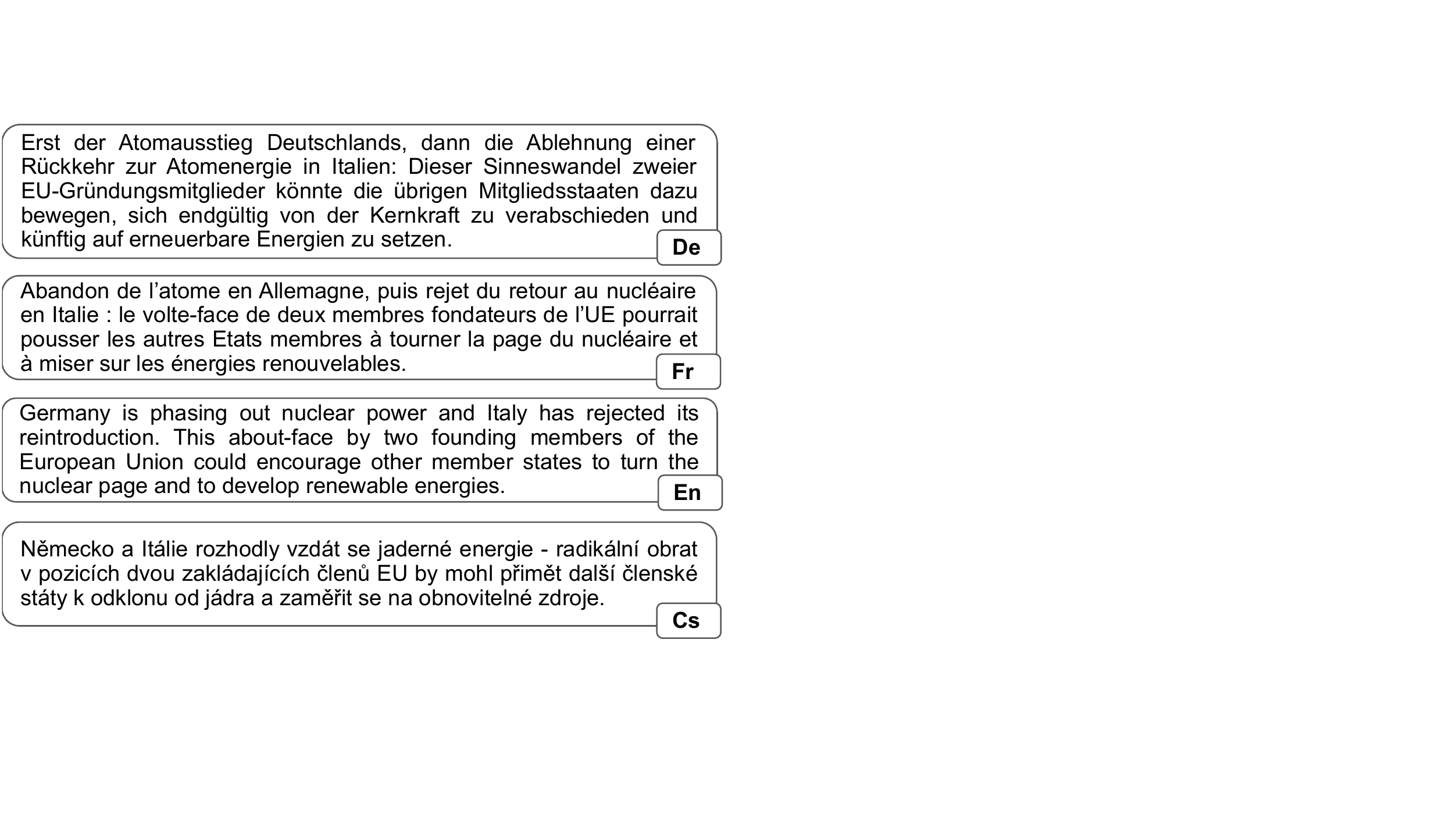}}
  \end{tabular}
\vspace*{-.5cm}
\caption{Example from Voxeurop dataset: source document in French and target
  summaries in German, French, English, and Czech.}
 \label{fig:clads-example-xnews}
 \end{figure*}

\section{Experiments}
\label{sec:appendix:experiments}

All our models were built on top of the fairseq library
\cite{ott2019fairseq} code base.

\paragraph{Text Processing}
For sentence splitting and tokenisation in German, French and English,
we used the Stanza Python NLP Package \cite{qi2020stanza}. For Czech,
we used the MorphoDiTa package \cite{11234/1-1836}.

\paragraph{Training Details}
For m\textsc{Bart50} \cite{tang2020multilingual}, we used the
checkpoint provided as \texttt{mMBART~50~finetuned~many-to-many} and
for m\textsc{Bart} the \texttt{mBART.cc25} checkpoint, both available
in the fairseq library \cite{ott2019fairseq}.  We reused
m\textsc{Bart}'s 250K sentencepiece
\cite{kudo-richardson-2018-sentencepiece} model which was trained
using monolingual data for 100 languages.  However, to reduce the size
of the model to fit our GPU availability we carried out the following
modifications.  We trimmed the vocabulary to 135K. We first applied
the sentencepiece encoder to the language sets in our XWikis corpus
(Table~\ref{tab:xwikis-pairs}) and the English data (used to train the
monolingual summariser ${\cal{D}}_{en \rightarrow en}$) to generate a
reduced dictionary.  Then, we trimmed the dictionary and the models'
embeddings (taking care to map indices from the original dictionary to
the reduced one).  We further slimmed-down the position embeddings
layer from 1,024 to 600.

Supervised fine-tuning of m\textsc{Bart} and m\textsc{Bart50} was
carried out for 20K updates with a batch size of 80 instances,
following previous work \cite{lewis-etal-2020-bart,mbart}.  We used
the Adam optimizer ($\epsilon$=1e-6 and $\beta_2$=0.98) with linear
learning rate decay scheduling. We set dropout rate to 0.3 and
attention-dropout to 0.1.  We used half precision (fp16) and
additionally set the weight decay to 0.01 and clipped gradients norm
to 0.1.  We fine-tuned with label smoothing and $\alpha$=0.2.  When
fine-tuning on English mono-lingual summarisation, we freeze the
embedding layer for m\textsc{Bart50} as it showed better zero-shot
results (but not for m\textsc{Bart} as zero shot results were not
improved).  We used 4 GPUs with 12GB of memory, fine-tuning took
2~days of training.

For the few-shot adaptation, we kept similar hyperparameters, except
that we used a much smaller batch size, i.e.,~8 instances, and ran 1K
updates (300 few-shot) and 5k (1k few-shot).  We monitored validation
perplexity and obtained checkpoints with best perplexity. All few-shot
variants used 1 GPU with 12GB of memory (and needed less than 10 hours
of training).  For the \textit{Few} approaches, we sampled a subset of
English $S_{en \rightarrow en}$ instances of  size similar to the
support set $S_{X \rightarrow en}$ of the adaptation task ${\cal T}_{X
  \rightarrow en}$ and doubled its size when in addition applying
CVT. The sample of unlabelled CVT instances had also 
 size similar to the task support set.  Adding more unlabelled data for
CVT hurts performance.  We combined data from the three tasks, English
monolingual, Few cross-lingual instances (task support set) and
unlabelled cross-lingual instances. We computed a weighted loss with
weights 0.5, 1 and 0.1 respectively (note that variants with no CVT
have 0 in the third weight).

We followed the same instance formatting as used in \citet{mbart}.  We
use special language ID tokens~\verb|<LID>|, postpend sentences with
the~\verb|</S>|, and prepend~\verb|<S>| at the beginning of each
sequence.

\section{Results}
\label{sec:appendix:results}

\paragraph{Full Set of Metrics and Results}
Tables~\ref{tab:XWikisSupervisedTest} and~\ref{tab:XNews} are the
extended versions of Tables~\ref{tab:XWikisSupervisedTestROUGEL}
and~\ref{tab:XNewsROUGEL} in the paper. Here, we report ROUGE-1/2/L F1
metrics.

\begin{table*}[t]
\centering
{\footnotesize
  \begin{tabular}{c@{\hspace{0.2cm}}c@{\hspace{0.2cm}}m{3.5cm}@{\hspace{0.5cm}}c@{\hspace{6pt}}c@{\hspace{6pt}}
  c@{\hspace{6pt}}c@{\hspace{6pt}}c@{\hspace{6pt}}c@{}}
 
\thickhline
   && & en & de-en & fr-en  & cs-en  \\
\thickhline

 \multicolumn{3}{c}{\textsc{Ext-Oracle}} & 36.40/14.21/31.33 & 27.78/\hspace{0.5em}5.51/23.75 & 29.37/\hspace{0.5em}6.73/25.01 & 29.33/\hspace{0.5em}5.99/25.09 \\
 \multicolumn{3}{c}{\textsc{Lead}} & 29.99/\hspace{0.5em}6.07/25.45 & 29.25/\hspace{0.5em}5.58/24.95& 28.97/\hspace{0.5em}5.57/24.74 & 28.58/\hspace{0.5em}5.10/24.35 \\
 \multicolumn{3}{c}{\textsc{LexRank}} & 30.06/\hspace{0.5em}6.43/25.23 & 28.71/\hspace{0.5em}5.57/24.22 & 28.82/\hspace{0.5em}5.64/24.33 & 27.88/\hspace{0.5em}5.04/23.68 \\[1.2ex]
 
 \hdashline
  \parbox[t]{2mm}{\multirow{8}{*}{\rotatebox[origin=m]{90}{m\textsc{Bart}}}} 
 & \multicolumn{2}{l}{Supervised} & 35.57/13.23/31.62 & 35.88/13.14/32.37 & 35.76/12.86/32.18 & 36.43/13.58/32.84 \\[1ex] % decode >> all-en=mlb150.ml50.lenpen2 en=mlb150.ml50.lenpen1
 & \multicolumn{2}{l}{Translated} & --- & 34.64/11.71/30.69 & 34.56/11.46/30.63 & 34.22/11.23/30.39 \\[1ex]
 & \multicolumn{2}{l}{Zero} & --- & 33.45/11.22/30.10 & 33.15/10.46/29.78 & 31.70/10.10/28.64 \\[1ex] %decode >> mlb150.ml50.lenpen2
 
  & \parbox[t]{2mm}{\multirow{4}{*}{\rotatebox[origin=m]{90}{Few}}}
 & 300 LF-MAML & --- & 34.32/11.53/30.84 & 33.87/10.96/30.44 & 33.58/11.04/30.15 \\
 && 300 FT  & ---    & 34.50/11.78/31.06 & 33.82/11.04/30.39 & 33.77/11.07/30.36 \\
 && 300 CVT & ---    & 33.81/11.36/30.40 & 33.57/10.82/30.12 & 32.63/10.61/29.39 \\
 && 1K LF-MAML & --- & 34.76/11.93/31.19 & 34.25/11.27/30.77 & 34.50/11.91/31.02 \\[1.2ex]
    
 \hdashline
 \parbox[t]{2mm}{\multirow{8}{*}{\rotatebox[origin=m]{90}{m\textsc{Bart50}}}} 
  & \multicolumn{2}{l}{Supervised}  & 36.60/13.73/32.53 & 36.64/13.96/32.95 & 35.59/12.70/31.84 & 37.56/14.57/33.72 \\[1ex]
  
  & \multicolumn{2}{l}{Translated} & --- & 35.49/12.46/31.53 & 35.30/12.33/31.35 & 35.15/12.07/31.25 \\[1ex]
  
  & \multicolumn{2}{l}{Zero} & --- & 35.37/12.32/31.70 & 34.66/11.49/30.97 & 34.73/11.83/31.14 \\[1ex]  %en: 36.63/13.69/32.56
 
  & \parbox[t]{2mm}{\multirow{4}{*}{\rotatebox[origin=m]{90}{Few}}}
  &  300 LF-MAML & --- & 35.61/12.74/31.96 &  34.82/11.86/31.17 & 35.35/12.32/31.73 \\
  && 300 FT  & --- & 35.45/12.45/31.77 & 35.01/12.04/31.39 &  35.43/12.30/31.67 \\
  && 300 CVT & --- & 35.45/12.41/31.77 & 34.77/11.53/31.08 &  35.53/12.52/31.91 \\
  && 1K LF-MAML  & --- & 35.69/12.73/32.01 &  35.09/12.08/31.46 & 35.63/12.65/32.00 \\
 
\thickhline
  \end{tabular}  
 }
 \caption{ROUGE-1/2/L F1 $X \rightarrow en$ XWikis test sets.}\label{tab:XWikisSupervisedTest}
\end{table*}

 \begin{table*}[t]
 \centering
 {\footnotesize
  \begin{tabular}{c@{\hspace{0.2cm}}c@{\hspace{0.2cm}}m{3.6cm}@{\hspace{0.2cm}}c@{\hspace{6pt}}c@{\hspace{6pt}}
  c@{\hspace{6pt}}c@{\hspace{6pt}}c@{\hspace{6pt}}c@{}}
  
 \thickhline
    &&& en & de-en & fr-en &  cs-en  \\
 \thickhline
     
 \multicolumn{3}{c}{\textsc{Ext-Oracle}} & 29.16/9.94/20.83 & 25.12/5.07/17.81 & 25.41/5.52/17.90 & 24.93/4.69/17.63 \\
 \multicolumn{3}{c}{\textsc{Lead}}       & 24.62/3.98/17.17 & 24.26/3.58/17.13 & 23.60/3.51/16.61 & 24.27/3.62/17.07 \\
 \multicolumn{3}{c}{\textsc{LexRank}}    & 24.22/3.59/16.65 & 23.20/3.09/16.32 & 23.32/3.21/16.32 & 23.32/3.16/16.48 \\[1.2ex]
 
 \hdashline

 \parbox[t]{2mm}{\multirow{3}{*}{\rotatebox[origin=m]{90}{m\textsc{Bart}}}}
&\multicolumn{2}{l}{Zero}  & 26.72/5.13/21.68 & 23.16/3.77/19.54 & 23.30/3.75/19.49 & 22.43/3.41/18.92 \\[1ex]
& \parbox[t]{2mm}{\multirow{2}{*}{\rotatebox[origin=m]{90}{Few}}}
& 300 LF-MAML & --- & 27.55/4.76/22.32 & 27.68/4.71/22.42 & 27.42/4.64/22.26 \\
&& 300 FT & --- & 26.62/4.60/21.86 & 26.78/4.48/21.74 & 26.42/4.57/21.72 \\[1ex]

\parbox[t]{2mm}{\multirow{3}{*}{\rotatebox[origin=m]{90}{m\textsc{Bart50}}}}
&\multicolumn{2}{l}{Zero} & 26.07/5.00/21.28 & 25.32/4.71/21.04 & 25.02/4.60/20.66 & 25.81/4.68/21.30 \\[1ex]
& \parbox[t]{2mm}{\multirow{2}{*}{\rotatebox[origin=m]{90}{Few}}}
&  300 LF-MAML & --- & 27.44/4.87/21.87 & 27.36/4.83/21.90 & 27.41/4.82/22.11 \\
&& 300 FT & --- & 27.13/4.76/21.79 & 27.01/4.61/21.53 & 27.35/4.79/21.95 \\[1.2ex]

 \thickhline
   \end{tabular}  
  }
  \caption{ROUGE-1/2/L F1 $X \rightarrow en$ Voxeurop test sets.}\label{tab:XNews}
 \end{table*}

\paragraph{Example Outputs}
Tables~\ref{tab:exDeEn},~\ref{tab:exFrEn},~and~\ref{tab:exCsEn} show
example outputs by m\textsc{Bart50} model variants for the three
language pairs German-English, French-English, and Czech-English,
respectively. Table~\ref{tab:exVoxeurop} shows example outputs
for the different m\textsc{Bart50} model variants on the Voxeurop
dataset.

\begin{table*}[t]
{\footnotesize
 \centering
 \begin{tabular}{cm{15cm}}  \thickhline
 \multirow{2}{*}[0.7ex]{\rotatebox[origin=l]{90}{Reference}} & 
 The Maria Canals International Music Competition (, ) is a music
 competition held yearly in the Palau de la M\'{u}sica Catalana,
 Barcelona. It was founded as a piano competition in 1954, but in 1964
 it was expanded so other modalities could be held occasionally. \\ 
 &  \\[1.7ex]
 
 \hline
 \parbox[t]{2mm}{\multirow{2}{*}[1.2ex]{\rotatebox[origin=c]{90}{Supervised}}} & 
 The ``Maria Canals International Music Competition'' () is an
 international music competition in Spain. It was founded in 1954 by
 Maria Canals i Cendr\'{o}s (1913--2010) and her husband Rossend Llates
 i Serrat (1899--1973). In 1964, the competition was expanded to
 include more than 120 competitions, with over 8.000 participants from
 100 countries. \\ 
 &  \\[1.5ex]
 
 \hline
 \parbox[t]{2mm}{\multirow{2}{*}[1.2ex]{\rotatebox[origin=c]{90}{Translated}}} & 
 The Maria Canals International Music Competition is an international
 music competition organised by the Spanish Ministry of the
 Interior. It is a member of the World Federation of International
 Music competitions. The competition is held every year from 1954 to
 2019.\\ 
&   \\[1.5ex]
 
 \hline
 \parbox[t]{2mm}{\multirow{2}{*}[1ex]{\rotatebox[origin=c]{90}{Zero}}} & 
 The Maria Canals International Music Competition is a
 multi-disciplinary music competition organised by the Spanish
 Ministry of the Interior. The competition is held annually in
 Spain. It was founded in 1954 by the Catalan pianist Maria canals i
 Cendr\'{o}s and her husband, Rossend Llates i Serrat.\\ 
 &   \\
 
 \hline
 \parbox[t]{2mm}{\multirow{2}{*}[2.2ex]{\rotatebox[origin=c]{90}{300 {\scriptsize LF-MAML}}}} & 
 The Maria Canals International Music Competition is a music
 competition held annually in Catalonia, Spain. The competition was
 founded in 1954 by the Catalan pianist Maria canals i Cendr\'{o}s and her
 husband, the composer Rossend Llates i Serrat (1899–1973). \\  
 & \\[2.4ex]
 
 \thickhline
 \end{tabular}
}
 \caption{Example with m\textsc{Bart50} based models outputs from the validation set for German-to-English.}\label{tab:exDeEn}
\end{table*}

\begin{table*}[t]
{\footnotesize
 \centering
 \begin{tabular}{cm{15cm}}
 \thickhline

 \multirow{2}{*}[-2ex]{\rotatebox[origin=l]{90}{Reference}} & \\
 &``Never Smile at a Crocodile'' is a comic song with music by Frank
 Churchill and lyrics by Jack Lawrence. The music, without the lyrics,
 was first heard in the Walt Disney animated film ``Peter
 Pan''. Following the film's release in 1953, the sung version with
 Lawrence's lyrics went on to become a children's song
 classic.\\[1.5ex] 
 
 \hline
 \parbox[t]{2mm}{\multirow{2}{*}[-1ex]{\rotatebox[origin=c]{90}{Supervised}}} & \\
 &``Never Smile at a Crocodile'' is a song by American
 singer-songwriter Charles Churchill. It was first released on the
 Disney soundtrack to the 1937 Disney film ``Blanche-Neige et les Sept
 Nains'' in 1939, when ``Peter Pan'' was already in the planning
 stages. However, Disney decided to discontinue the recording until
 1949. \\[1.5ex] 
 
 \hline
 \parbox[t]{2mm}{\multirow{2}{*}[-2ex]{\rotatebox[origin=c]{90}{Translated}}} & \\
& ``Never Smile at a Crocodile'' is a song from the 1937 Disney film
``Snow White and the Seven Dwarfs''. It was composed by Charles
Churchill, who had composed most of the soundtrack for the 1937 film,
``Peter Pan''. However, the original lyrics were later recorded by
several singers, including Jerry Lewis and Rolf Harris. The song
became one of the top ten best-selling children's songs that year.
\\[1.5ex] 
 
 \hline
 \parbox[t]{2mm}{\multirow{2}{*}[-2.5ex]{\rotatebox[origin=c]{90}{Zero}}} &  \\
 & ``Never Smile at a Crocodile'' is a song by American singer-songwriter John Churchill. It was first released in 1953 as the soundtrack to the 1937 Disney film, ``Blanche-Neige and the Seven Nains''. The song was later re-released as a CD in 1997.  \\[1.3ex]
 
 \hline
 \parbox[t]{2mm}{\multirow{2}{*}[-2ex]{\rotatebox[origin=c]{90}{ 300 FT}}} &  \\
 & ``Never Smile at a Crocodile'' is a song by American singer-songwriter John Churchill. It was first released in 1939 as the soundtrack to the 1937 Disney film ``Blanche-Neige and the Seven Nains''. The song was later re-released as the lead single from the 1953 film ``Peter Pan''.\\[1.3ex]
 
 \thickhline
 \end{tabular}
}
 \caption{Example with m\textsc{Bart50} based models outputs from the
   validation set for French-to-English.}\label{tab:exFrEn} 
\end{table*}

\begin{table*}[t]
{\footnotesize
 \centering
 \begin{tabular}{cm{15cm}}
 \thickhline

 \multirow{2}{*}[1.9ex]{\rotatebox[origin=l]{90}{Reference}} &
 Privacy International (PI) is a UK-based registered charity that defends and promotes the right to privacy across the world. First formed in 1990, registered as a non-profit company in 2002 and as a charity in 2012, PI is based in London. Its current executive director, since 2012, is Dr Gus Hosein.\\
 &  \\
 
 \hline
 \parbox[t]{2mm}{\multirow{2}{*}[2ex]{\rotatebox[origin=c]{90}{Supervised}}} &
 Privacy International is an international non-governmental organization dedicated to protecting the privacy of individuals, companies, and organizations. It is the world's largest private security organization. The organization is headquartered in Washington, D.C., and operates under the auspices of the Worldwide Interbank Financial Telecommunication (SWIFT).\\
 &  \\[1.5ex]
 
 \hline
 \parbox[t]{2mm}{\multirow{2}{*}[2ex]{\rotatebox[origin=c]{90}{Translated}}} &
 The Society for Worldwide Interbank Financial Telecommunication (SWIFT) is a non-profit organization dedicated to providing financial information to customers of financial institutions under SWIFT. The organization was founded in 1997 and is currently headquartered in Los Angeles, California. \\
 & \\[1.4ex]
 
 \hline
 \parbox[t]{2mm}{\multirow{2}{*}[1.4ex]{\rotatebox[origin=c]{90}{Zero}}} &
 Privacy International is a non-profit organization dedicated to protecting the privacy and privacy of financial institutions. It was founded in 1997 and has been active since 2003. It is the largest private security organization in the United States. The organization is a member of the World Bank and the International Monetary Fund.\\
 &   \\
 
 \hline
 \parbox[t]{2mm}{\multirow{2}{*}[2ex]{\rotatebox[origin=c]{90}{1k {\scriptsize LF-MAML}}}} &
 Privacy International is a non-profit organization dedicated to
 protecting and protecting the privacy of individuals, companies and
 corporations. It was founded in 1997 and is one of the largest
 private sector organizations in the United States. The organization's
 mission is to protect and protect the privacy and data of individuals
 and companies \\ 
 & \\[1.6ex]
 
 \thickhline
 \end{tabular}
}
 \caption{Example with m\textsc{Bart50} based models outputs from the validation set for Czech-to-English.}\label{tab:exCsEn}
\end{table*}

\begin{table*}[t]
{\footnotesize
 \centering
 \begin{tabular}{ccm{14.5cm}}
 \thickhline

 &\multirow{2}{*}[-2ex]{\rotatebox[origin=l]{90}{Gold}} & \\
 && One in every five young Europeans is out of a job, and even one in two in some countries. Numbers like these were enough to have the young generation rebel against governments in the Arab world, remarks a Polish columnist. What will happen if our social model deprives young people of all hope? \\[1.5ex]
 
  \hline
  \parbox[t]{2mm}{\multirow{2}{*}[-3ex]{\rotatebox[origin=c]{90}{en}}}
 &\parbox[t]{2mm}{\multirow{2}{*}[-1ex]{\rotatebox[origin=c]{90}{\textsc{Oracle}}}} & \\
 && For many international education experts, a university education – bachelor or master’s degree, doctorate – is the measure of all things. And it is true that the time-frame may not be ideal, as the German system is strongly dependent on the economy. \\[1.5ex]

  \hline
  \parbox[t]{2mm}{\multirow{2}{*}[-3ex]{\rotatebox[origin=c]{90}{en}}}
 &\parbox[t]{2mm}{\multirow{2}{*}[-2ex]{\rotatebox[origin=c]{90}{\textsc{Lead}}}} & \\
 && More than 5.5m young Europeans are without jobs. In the crisis countries in southern Europe, a generation is coming of age with few prospects: one in two Spaniards and Greeks under 25 are unemployed, and it's one in three in Italy and Portugal . To them, Germany must \\[1.5ex]
 
  \hline
  \parbox[t]{2mm}{\multirow{2}{*}[-3ex]{\rotatebox[origin=c]{90}{en}}}
 &\parbox[t]{2mm}{\multirow{2}{*}[-1ex]{\rotatebox[origin=c]{90}{\textsc{LexRank}}}} & \\
 && As do young southern Europeans who are leaving home to come to Germany to find a job or receive vocational training. They not only lack companies willing to create apprenticeship positions, and patient “masters” happy to pass on their know-how to “their” apprentices, but also the institutions, and \\[1.5ex]
 
  \hline
 \parbox[t]{2mm}{\multirow{4}{*}[-1ex]{\rotatebox[origin=c]{90}{en}}}
 &\parbox[t]{2mm}{\multirow{2}{*}[-3ex]{\rotatebox[origin=c]{90}{Zero}}} & \\
 && Youth unemployment in Europe has risen to 52\% in Spain and Greece. In countries such as the United Kingdom, the jobs that are on offer are invariably short-term contracts. Precarious work is now the only option for a generation threatened by employment and poverty. However, in Europe, we may not have dictators to depose, but Monti’s remarks are an indirect admission of the capitulation of democracy in response to the crisis. \\[1.5ex]

 \hline
 \parbox[t]{2mm}{\multirow{4}{*}[-1ex]{\rotatebox[origin=c]{90}{de-en}}}
 &\parbox[t]{2mm}{\multirow{2}{*}[-3ex]{\rotatebox[origin=c]{90}{Zero}}} & \\
 && This article is a list of the events that have taken place in Greece, Italy, Spain, and the United Kingdom in the last decades of the twentieth century. The events that took place in Italy, Greece, Spain and Italy in the first decade of the twenty-first century have been described as "the most important events in the history of the European Union". \\[1.5ex]
 
 \cline{2-3}
 &\parbox[t]{2mm}{\multirow{2}{*}[-3ex]{\rotatebox[origin=c]{90}{Few}}} & \\
 && In Europe, youth unemployment is on the rise. In Spain and Greece, it is rising to 52 per cent. But what will happen if the governments of Greece, Spain and Italy stop cutting their pensions? \\[3ex]
 
  \hline
  \parbox[t]{2mm}{\multirow{4}{*}[-1ex]{\rotatebox[origin=c]{90}{fr-en}}}
 &\parbox[t]{2mm}{\multirow{2}{*}[-3ex]{\rotatebox[origin=c]{90}{Zero}}} & \\
 && This is a list of events that have taken place in the last decades of the twentieth century in Europe. The most recent events in the history of the European Union have been the events in Greece, Spain, Spain and the United Kingdom. \\[3ex]
 
 \cline{2-3}
 &\parbox[t]{2mm}{\multirow{2}{*}[-3ex]{\rotatebox[origin=c]{90}{Few}}} & \\
 && A message of hope for young people in Europe has been delivered by Italian Prime Minister Mario Monti, who has deplored the fact that the unemployment rate of 20 years old is now a lost generation. But what will happen when they are no longer in the world or when the governments of Greece, Spain and Italy reduce the level of pensions? \\[1.5ex]
 
  \hline
  \parbox[t]{2mm}{\multirow{4}{*}[-1ex]{\rotatebox[origin=c]{90}{cs-en}}}
 &\parbox[t]{2mm}{\multirow{2}{*}[-3ex]{\rotatebox[origin=c]{90}{Zero}}} & \\
 && This is a list of events that have taken place in the European Union in the past two decades. This list includes the events that occurred in the last decade of the twentieth century, including the events of the Arab revolutions, the collapse of the European social model, and the fall in the living standards of young people \\[1.5ex]
 
 \cline{2-3}
 &\parbox[t]{2mm}{\multirow{2}{*}[-3ex]{\rotatebox[origin=c]{90}{Few}}} & \\
 && Whatever leaders do this week, they are not going to bridge the gap between unemployment in Europe and poverty in the Middle East. Instead, young people should take to the streets in Brussels to express their support for Europe, argues Mario Monti. \\[2ex]
 
 \thickhline
 \end{tabular}
}
 \caption{Examples from Voxeurop datasets. We show Gold summary together with three extractive
 baselines (\textsc{Ext-Oracle}, \textsc{Lead} and \textsc{LexRank}) on the input English document for comparison. For each cross lingual task
 (de-en, fr-en, and cs-en), we report \textsc{Bart50} Zero and Few Shot FL-MAML variants.}\label{tab:exVoxeurop}
\end{table*}

\end{document}